\documentclass[runningheads]{llncs}
\usepackage{graphicx}
\usepackage{tikz}
\usepackage{comment}
\usepackage{amsmath,amssymb} % define this before the line numbering.
\usepackage{color}

\usepackage[accsupp]{axessibility}  % Improves PDF readability for those with disabilities.

\usepackage{ragged2e}
\usepackage{graphicx}
\usepackage{amsmath}
\usepackage{amssymb}
\usepackage{booktabs}
\usepackage{enumitem}
\usepackage{caption}

\usepackage{times}
\usepackage{epsfig}

\usepackage[pagebackref,breaklinks,colorlinks]{hyperref}

\usepackage{amsfonts}
\usepackage{amsmath}

\usepackage{etoolbox} 
\newcommand{\MyMapTemplatePrefixc}[4]{\expandafter#1\csname#3#4\endcsname{#2{#4}}} % it remembles a template: \#3#4 --> #2{#4}
\forcsvlist{\MyMapTemplatePrefixc {\def} {\mathcal}{c}} {A,B,C,D,E,F,G,H,I,J,K,L,M,N,O,P,Q,R,S,T,U,V,W,X,Y,Z}  % e.g., \cA --> \mathcal{A}

\newcommand{\MyMapTemplatePrefixtb}[5]{\expandafter#1\csname#4#5\endcsname{#2{#3{#5}}}} % it remembles a template: \#3#4 --> #2{#4}
\forcsvlist{\MyMapTemplatePrefixtb {\def} {\tilde}{\mathbf}{t}} {A,B,C,D,E,F,G,H,I,J,K,L,M,N,O,P,Q,R,S,T,U,V,W,X,Y,Z}  % e.g., \tA --> \tilde{A}
\forcsvlist{\MyMapTemplatePrefixtb {\def} {\tilde}{\mathbf}{t}} {0,1,a,b,c,d,e,f,g,h,i,j,k,l,m,n,o,p,q,r,s,u,v,w,x,y,z}  % e.g., \ta --> \tilde{a}

\newcommand{\MyMapTemplateNoPrefix}[3]{\expandafter#1\csname#3\endcsname{#2{#3}}}
\forcsvlist{\MyMapTemplateNoPrefix {\def} {\mathbf} } {0,1,a,b,c,d,e, f, g, h, i, j, k, l, m, n, o, p, q, r, s, t, u, v, w, x, y, z} % e.g., \a --> \mathbf{a}
\forcsvlist{\MyMapTemplateNoPrefix {\def} {\mathbf} } {A,B,C,D,E,F,G,H,I,J,K,L,M,N,O,P,Q,R,S,T,U,V,W,X,Y,Z}  % e.g., \A --> \mathcal{A}

\def\resp{\emph{resp.}}

\usepackage{xspace}
\usepackage[marginal]{footmisc}
\usepackage[misc]{ifsym}

\makeatletter
\DeclareRobustCommand\onedot{\futurelet\@let@token\@onedot}
\def\@onedot{\ifx\@let@token.\else.\null\fi\xspace}

\def\eg{\emph{e.g}\onedot} 
\def\ie{\emph{i.e}\onedot} 
\def\cf{\emph{cf}\onedot} 
\def\etc{\emph{etc}\onedot} \def\vs{\emph{vs}\onedot}
 
\def\etal{\emph{et al}\onedot}
\makeatother

\usepackage[capitalize]{cleveref}
\crefname{section}{Sec.}{Secs.}
\Crefname{section}{Section}{Sections}
\Crefname{table}{Table}{Tables}
\crefname{table}{Tab.}{Tabs.}

\begin{document}
% \renewcommand\thelinenumber{\color[rgb]{0.2,0.5,0.8}\normalfont\sffamily\scriptsize\arabic{linenumber}\color[rgb]{0,0,0}}
% \renewcommand\makeLineNumber {\hss\thelinenumber\ \hspace{6mm} \rlap{\hskip\textwidth\ \hspace{6.5mm}\thelinenumber}}
% \linenumbers
\pagestyle{headings}
\mainmatter
\def\ECCVSubNumber{4761}  % Insert your submission number here

\title{Motion and Appearance Adaptation for Cross-Domain Motion Transfer} % Replace with your title

% INITIAL SUBMISSION 
%\begin{comment}
% \titlerunning{ECCV-22 submission ID \ECCVSubNumber} 
% \authorrunning{ECCV-22 submission ID \ECCVSubNumber} 
% \author{Anonymous ECCV submission}
% \institute{Paper ID \ECCVSubNumber}
%\end{comment}
%******************

% CAMERA READY SUBMISSION
% \begin{comment}
\titlerunning{Motion and Appearance Adaptation for Cross-Domain Motion Transfer}
% If the paper title is too long for the running head, you can set
% an abbreviated paper title here
%
\author{Borun Xu\inst{1} \and
Biao Wang\inst{2} \and
Jinhong Deng\inst{1} \and
Jiale Tao\inst{1} \and
Tiezheng Ge\inst{2} \and
Yuning Jiang\inst{2} \and
Wen Li\inst{1}\thanks{The corresponding author} \and
Lixin Duan\inst{1}}

\authorrunning{Xu et al.}
% First names are abbreviated in the running head.
% If there are more than two authors, 'et al.' is used.
%

\institute{University of Electronic Science and Technology of China \and
Alibaba Group\\
\email{xbr\_2017@std.uestc.edu.cn,\\
\{jhdeng1997, jialetao.std, liwenbnu, lxduan\}@gmail.com,\\
\{eric.wb, tiezheng.gtz, mengzhu.jyn\}@alibaba-inc.com}}

% \institute{University of Electronic Science and Technology of China \and
% \email{@springer.com}
% Alibaba Group\\
% \email{\{abc,lncs\}@uni-heidelberg.de}}
% \end{comment}
%******************
\maketitle

\begin{abstract}
Motion transfer aims to transfer the motion of a driving video to a source image. When there are considerable differences between object in the driving video and that in the source image, traditional single domain motion transfer approaches often produce notable artifacts; for example, the synthesized image may fail to preserve the human shape of the source image (\cf~Fig.~\ref{fig:figure-1} (a)). To address this issue, in this work, we propose a Motion and Appearance Adaptation (MAA) approach for cross-domain motion transfer, in which we regularize the object in the synthesized image to capture the motion of the object in the driving frame, while still preserving the shape and appearance of the object in the source image. On one hand, considering the object shapes of the synthesized image and the driving frame might be different, we design a shape-invariant motion adaptation module that enforces the consistency of the angles of object parts in two images to capture the motion information. On the other hand, we introduce a structure-guided appearance consistency module designed to regularize the similarity between the corresponding patches of the synthesized image and the source image without affecting the learned motion in the synthesized image. Our proposed MAA model can be trained in an end-to-end manner with a cyclic reconstruction loss, and ultimately produces a satisfactory motion transfer result (\cf~Fig.~\ref{fig:figure-1} (b)). We conduct extensive experiments on human dancing dataset Mixamo-Video to Fashion-Video and human face dataset Vox-Celeb to Cufs; on both of these, our MAA model outperforms existing methods both quantitatively and qualitatively.

% \keywords{motion transfer, domain adaptation, video synthesis}
\end{abstract}

%%%%%%%%% BODY TEXT
\section{Introduction}
\label{sec:Introduction}

%-------------------------------------------------------------------------

\begin{figure}[t]
  \centering
  \includegraphics[width=0.9\linewidth]{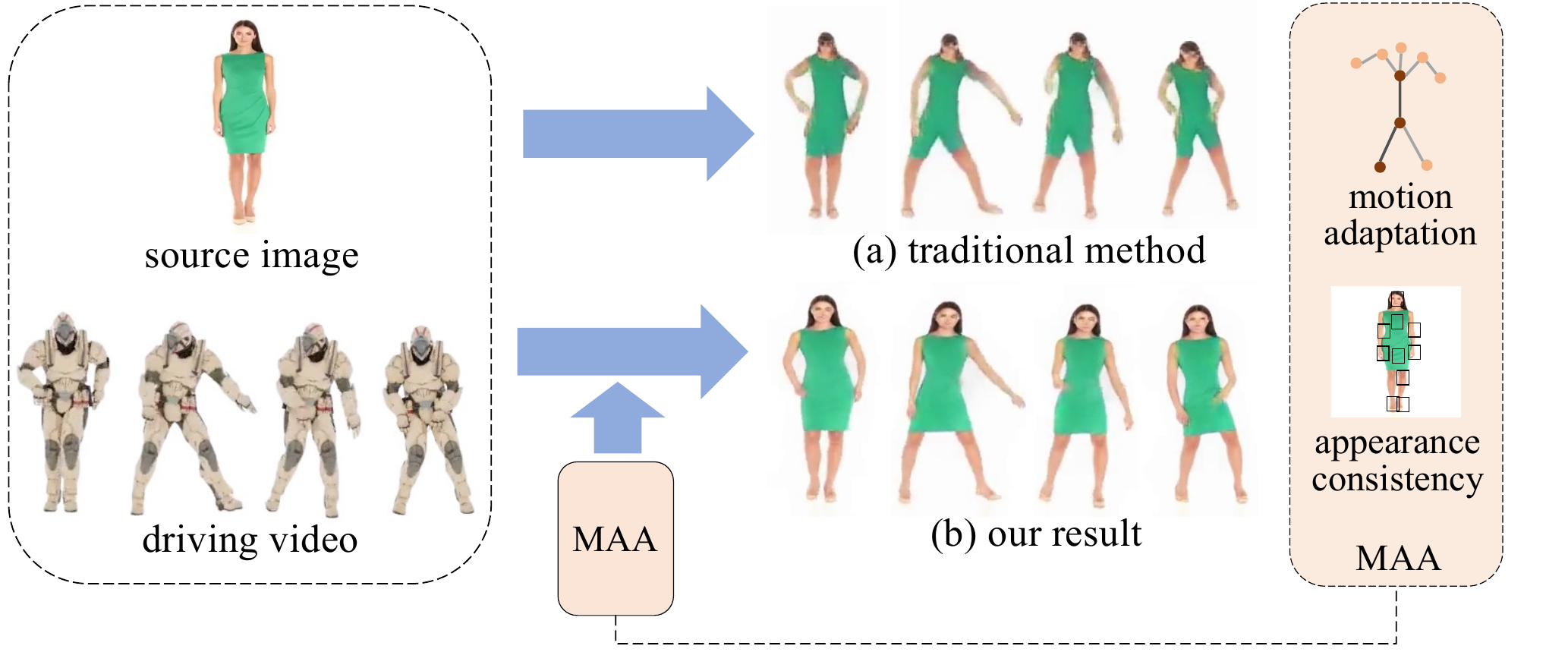}
%   \wenli{It is better to change the drving frame to driving video, by stacking several frames.}
   \caption{ Motion transfer results. (a) is generated by traditional motion transfer model trained on source domain videos only and (b) is generated by our proposed MAA model}
   \label{fig:figure-1}
  % \vspace{-15pt}
\end{figure}

%------------------------------------------------------------------------

Given a source image and a driving video of the same object, motion transfer~(\textit{a.k.a.} image animation) aims to generate a synthesized video that mimics the motion of the driving video while preserving the appearance of the source image. It recently received increasing attention, due to its potential applications in real-world scenarios, such as face swapping\cite{wiles2018x2face,siarohin2019first,xu2021move}, dance transferring\cite{chan2019everybody}, \etc. 

% arbitrary object
Many works in this field focus on the single-domain motion transfer \cite{siarohin2019animating,siarohin2019first,siarohin2021motion}, where the driving video and source image come from the same domain. However, in real applications, there are often requirements to transfer motion among different domains. For example, as shown in Fig~\ref{fig:figure-1}, the e-commerce companies might be interested in animating a fashion model to attract consumers by learning the robot dance from a Mixamo character. However, due to the differences in shape and cloth between the Mixamo character and the fashion model, traditional single-domain motion transfer approaches often produce notable artifacts in the synthesized image (\eg, failing to preserve the human shape of the fashion model (\cf~Fig.~\ref{fig:figure-1} (a))). 

% \footnotetext[1]{The corresponding author}

To this end, in the present work, we study the cross-domain motion transfer problem and propose a novel Motion and Appearance Adaptation (MAA) approach to address this issue. Specifically, traditional motion transfer methods usually take two arbitrary frames of the same video as source image and driving frame for learning motion with a reconstruction loss, because the two frames share the same appearance and shape. However, such training mode cannot be directly applied to the cross-domain motion transfer because no ground-truth is available. In our proposed MAA approach, we build a cyclic reconstruction pipeline inspired by CycleGAN\cite{zhu2017unpaired} and cross-identity\cite{jeon2020cross}. In particular, given a source image and a driving frame obtained from different domains, we first obtain a synthesized image using a basic motion transfer (MT) model, \eg, the model in\cite{siarohin2019first} or\cite{siarohin2021motion}. We next arbitrarily take another frame from the driving video as the source image and the synthesized image as a driving frame, and input them into the basic MT model to produce the second synthesized image. Because the second synthesized image should mimic both the motion and appearance of original driving frame, a cyclic reconstruction loss can be applied for training. In this way, we obtain a motion transfer model for cross-domain motion transfer. 

Moreover, since the source image and driving frame are drawn from different domains, while the topology of the object structure (\eg, the skeleton) is similar, the configurations of the object structure (\eg, the human body shape) often deviate. When doing motion transfer, we should be aware of such difference and keep the object shape of synthesized image be similar to the source image while unaffected by the driving frame. For this purpose, we design a shape-invariant motion adaptation module and a structure-guided appearance consistency module to regularize the basic motion transfer model. 

Specifically, in the shape-invariant motion adaptation module, we design an angle consistency loss to enforce the angles of the corresponding object parts in the synthesized image to be similar to those of the driving frame, such that the motion of this frame can be mimicked well without changing the object shape. In the structure-guided appearance consistency module, we extract image patches from the synthesized image and the source images based on the object structure and enforce the corresponding patches to be similar; this ensures that the appearance of the synthesized image and the source image are consistent, even though the motions of the two images are different. 

The entire process can be trained in an end-to-end manner, and finally our MAA model can effectively perform motion transfer across domains while also properly preserving the shape and appearance of the object (\cf Fig.~\ref{fig:figure-1} (b)). 
% Our method makes arbitrary object motion transfer model adaptive to the target domain, e.g. testing data and finally get better results as we demonstrate in the figure (b) of \cref{fig:figure-1}. 
We validate our proposed approach on two pairs of datasets: the human body datasets Mixamo-Video to Fashion-Video\cite{zablotskaia2019dwnet} and the human face datasets Vox-Celeb\cite{nagrani2017voxceleb} to Cufs\cite{wang2008face}. Extensive experimental results demonstrate the effectiveness of our proposed approach. Our source code will be released soon.

\section{Related Work}
\label{sec:Related Work}
\textbf{Motion Transfer}: Current motion transfer approaches can be categorized into two types: model-based and model-free approaches. The model-based approaches mainly focus on human body pose transfer\cite{ma2017pose,ma2018disentangled,balakrishnan2018synthesizing}, which utilize a pre-trained pose estimator or key point detector to extract the pose of driving image as a guidance information. And a number of researchers followed such setting\cite{liu2019liquid,ren2020deep,zhu2019progressive,li2019dense,huang2021few,kappel2021high}. Moreover, a series of works apply this model-based pattern on human facial expression transfer\cite{burkov2020neural,chen2020puppeteergan,gu2020flnet}. Like body pose transfer, they also employ a pre-trained facial landmark detector to model the facial expression. 

The model-free approaches\cite{siarohin2019animating,siarohin2019first,jeon2020cross,siarohin2021motion,tao2022structure} does not rely on pre-trained third-party models, and extend the model-based method to arbitrary objects. Aliaksandr \etal\cite{siarohin2019animating} proposed a model-free motion transfer model Monky-Net that can apply motion transfer on arbitrary objects with an unsupervised key point detector trained by reconstruction loss \cite{jakab2018unsupervised}. Aliaksandr \etal\cite{siarohin2019first} further improved Monkey-Net to FOMM to solve the large motion problem. The unsupervised key point detector is also utilized in FOMM, with local affine transformations being added for motion modeling. A generator module is utilized to generate final result with the warped source image feature. Subin \etal\cite{jeon2020cross} proposed pose attention mechanism with an unsupervised key point detector to model motion. Recently, Aliaksandr \etal\cite{siarohin2021motion} improved FOMM with an advanced motion model and background motion model to MRAA. Although promising results are achieved for the single domain motion transfer, these methods might suffer from performance degradation when the source image and driving video come from different domains, where a considerable appearance difference often exists. Recently, Wang \etal\cite{wang2020self} used encoder based motion transfer approach which can be applied to the cross-domain scenario, and better results are achieved compared with the single domain motion transfer Monkey-Net model. In contrast, our proposed MAA approach is a general framework, and can integrate traditional motion transfer model like FOMM and MRAA to produce excellent results for large motion.

\textbf{Domain Adaptation:} Many works have been proposed to handle the scenario where the training and test data comes from different domains for different computer vision tasks , \eg,  classification~\cite{DMCD}, semantic segmentation~\cite{Liu_2022_CVPR,Liu_2021_ICCV,9616392_Dong,What_Transferred_Dong_CVPR2020}, object detection~\cite{chen2018domain,deng2021unbiased}, pose estimation\cite{li2021synthetic,zhang2021keypoint}, \etc. A majority of works were developed to learn  domain-invariant features using the domain adversarial learning~\cite{tzeng2017adversarial,ganin2015unsupervised}. Cross-domain motion transfer is more complicated, since we need to capture motion from the the driving video while preserving the appearance from the source domain. Nevertheless, the strategies proposed in traditional domain adaptation works might be useful to help motion transfer. For example, we apply the cyclic training pipeline inspired by CycleGAN~\cite{zhu2017unpaired}, and build our patch-based appearance consistency module based on Patch-GAN~\cite{isola2017image}.

\section{Methodology}
\label{sec:Methodology}
In this section, we present our Motion and Appearance Adaptation approach for cross-domain motion transfer. Formally, let us denote a driving video as $V_d = \{I_d^{i}|_{i=1}^T\}$, where each $I_d^i$ is a driving frame, while a source image is denoted as $I_s$; thus, the task of motion transfer is to synthesize a new video $\hat{V}_d =  \{\hat{I}_d^{i}|_{i=1}^T\}$, where each $\hat{I}_d$ adequately captures the object motion in the corresponding driving frame $I_d^i$ while also preserving the object appearance of the source image $I_s$. 

% To illustrate the cross domain motion transfer problem. 
The appearance of an object roughly consists of two aspects, \emph{shape} and \emph{texture}. The shape largely refers to its geometric property (\eg, length, slimness, etc.), while the texture usually means how the object looks like regardless of its shape (\eg, dresses with different colors). Traditional motion transfer methods generally assume that the driving frame and the source image are derived from the same domain, where they implicitly suppose the object shapes are similar. Consequently, when the source image is derived from a new domain with different object shapes, these methods often fail to preserve the shape of the object in the source image. 

In this work, we study the cross-domain motion transfer problem, in which the source image and driving frame are from different domains. In other words, there might be considerable differences in appearance between them in terms of both shape and texture. An example is given in Fig.~\ref{fig:figure-1}, where both the clothes and body shapes of the fashion model and the Mixamo character exhibit notable differences. 

In what follows, we first present an overview of the pipeline of our proposed MAA approach in \cref{sec:overview}, after which we present the shape-invariant motion adaptation (SIMA) module and structure-guided appearance consistency~(SGAC) module in \cref{sec:SIMA} and \cref{sec:SGAC} respectively; these effectively learning the motion and appearance from the driving frame and source image, respectively.

%-------------------------------------------------------------------------

\begin{figure*}
  \centering
  \includegraphics[width=0.9\linewidth]{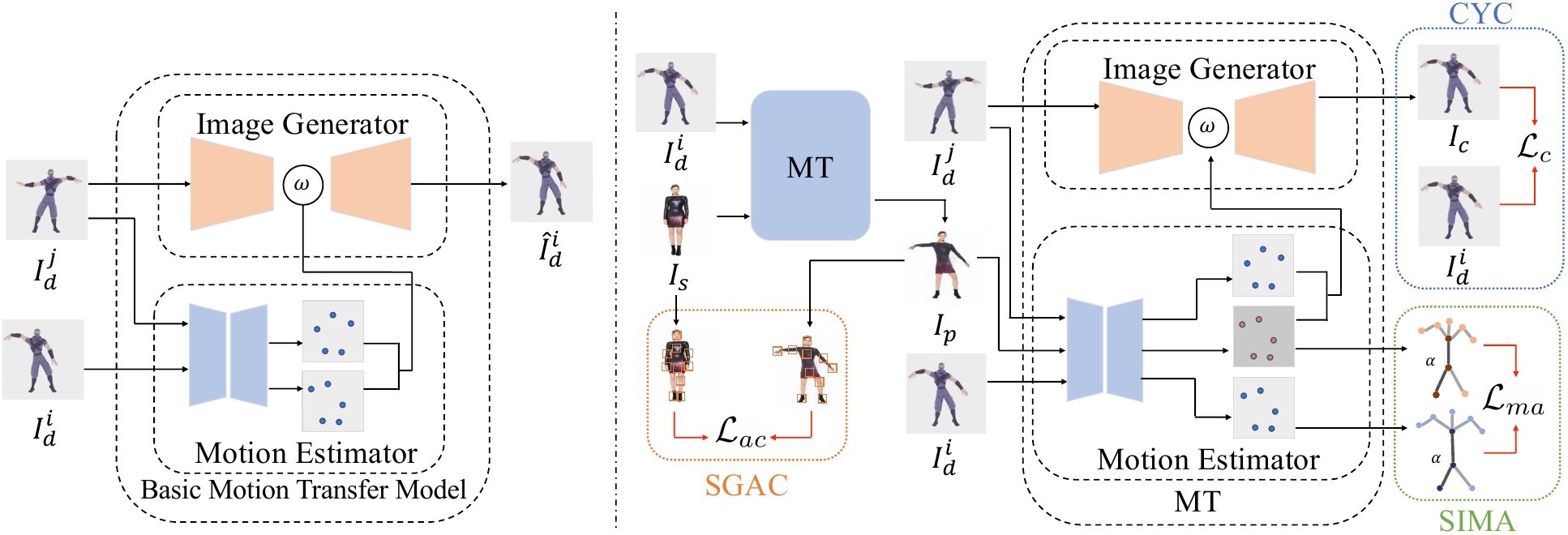}
  \caption{The pipeline of our proposed method. The left-hand side is the architecture of the traditional single domain motion transfer model FOMM\cite{siarohin2019first}, which is used as a basic motion transfer model in our approach. Moreover, the right-hand side is the framework of our proposed MAA method where we design a cyclic reconstruction loss (CYC), a shape-invariant motion adaptation (SIMA) and a structure-guided appearance consistency (SGAC) module}
  \label{fig:fig-pipeline}
  % \vspace{-10pt}
\end{figure*}

%------------------------------------------------------------------------

\subsection{Overview}
\label{sec:overview}
% \wenli{Have you checked the correctness of formulations?}

We design a cyclic training pipeline for cross-domain motion transfer, as shown in the right-hand part of \cref{fig:fig-pipeline}. The pipeline consists of a basic motion transfer model, our proposed shape-invariant motion adaptation module and structure-guided appearance consistency module, and a cyclic loss. 
% The basic motion transfer model follows the traditional motion transfer model, \eg, FOMM [xxx] or MCAA [xxx]. 

\textbf{Basic Motion Transfer Model:} The basic motion transfer (MT) model follows the traditional motion transfer model\cite{siarohin2019first,siarohin2021motion}. We here illustrate the basic MT model by taking FOMM\cite{siarohin2019first} as an example, and other models like\cite{siarohin2021motion} can be similarly integrated into our pipeline. 

As shown in the left-hand part of \cref{fig:fig-pipeline}, traditional motion transfer methods typically employ a reconstruction training mode for learning and synthesizing motion. During the training phase, they select two arbitrary frames from the driving video as the source image and driving frame, which are used as input of the MT model. For each image, the motion keypoints and their local affine transformation are extracted using a motion estimator, where the motion keypoints can be conceptualized as the centroids of moving object parts. The dense motion flow from the source image to the driving frame can therefore be estimated using their motion keypoints and affine transformations. In the next step, the dense motion flow is used to warp the feature map of the source image, and produce the synthesized image $\hat{I}_d^i$ using the image generator. A perceptual loss is used as the reconstruction loss after the image generator to ensure that the synthesized image $\hat{I}_d^i$ fully reconstructs the driving frame $I_d^i$ as in\cite{siarohin2019first}:
\begin{equation}
    \mathcal{L}_{r} = \sum_{l=k}^{K} | F_{l}(\hat{I}_d^i) - F_{l}(I_d^i)|
    \label{eq:reconstruction loss}
\end{equation}
where $ F_{l}(\cdot) $ is feature map output by the $l$-th layer of a pre-trained VGG-19 network\cite{simonyan2014very}. 

Researchers have proposed different method\cite{siarohin2021motion} to improve the motion estimator in order to more precisely extract motion information, yet the motion representation (\ie, motion keypoints and affine transformations) remains similar. In the interests of simplicity, we depict only the motion keypoints in \cref{fig:fig-pipeline}, which are related to our MAA approach. Readers can refer to\cite{siarohin2019first} for further details. 

\textbf{Cyclic Training Pipeline:} In cross-domain motion transfer, the source image and driving video are obtained from different domains. So it is undesirable to pick a frame in the driving video as a source image and apply the reconstruction loss after the image generator, as the model will inevitably be overfitted to the driving video, which will lead to artifacts in the synthesized image. 

To address this issue, we build a cyclic reconstruction framework inspired by the CycleGAN\cite{zhu2017unpaired} and cross-identity\cite{jeon2020cross}. As shown in the right-hand side of Fig~\ref{fig:fig-pipeline}, we employ two basic basic MT models that share the same parameters. Given a source image $I_s$ and a driving frame $I_d^{i}$, we first obtain a synthesized image $I_p$ using the basic MT model. Since there is no ground-truth for the synthesized image, the reconstruction loss cannot be used for $I_p$. 

We then take the synthesized image $I_p$ as a driving frame, along with an arbitrary frame  $I_d^{j}$ as the source image, and these are input again into the basic MT model to produce another synthesized image $I_c$. Intuitively, $I_c$ should mimic the motion of $I_p$, as well as $I_d^{i}$, since we expect $I_p$ to mimic the motion of $I_d^{i}$. At the same time, $I_c$ should also preserve the appearance of $I_d^{j}$, as well as $I_d^{i}$, which is derived from the same driving video as $I_d^{j}$. This allows us to employ $I_d^{i}$ and the cyclically generated $I_c$ to create a reconstruction loss for training. More specifically, we employ the perceptual loss 
similarly as in \cref{eq:reconstruction loss}:
\begin{equation}
    \mathcal{L}_{c} = \sum_{l=k}^{K} | F_{l}(I_c) - F_{l}(I_d^i)|
    \label{eq:cyclic reconstruction loss}
\end{equation}

While the cyclic reconstruction loss enables us to train the motion transfer model in the cross-domain setting, this is only a weak supervision that cannot fully guarantee a satisfactory result. We therefore further introduce the shape-invariant motion adaptation module and patch-based appearance model to regularize the motion transfer process, which will be explained in more detail below. 

\subsection{Shape-invariant Motion Adaptation}
\label{sec:SIMA}
%-------------------------------------------------------------------------

\begin{figure}
  \centering
  \includegraphics[width=0.5\linewidth]{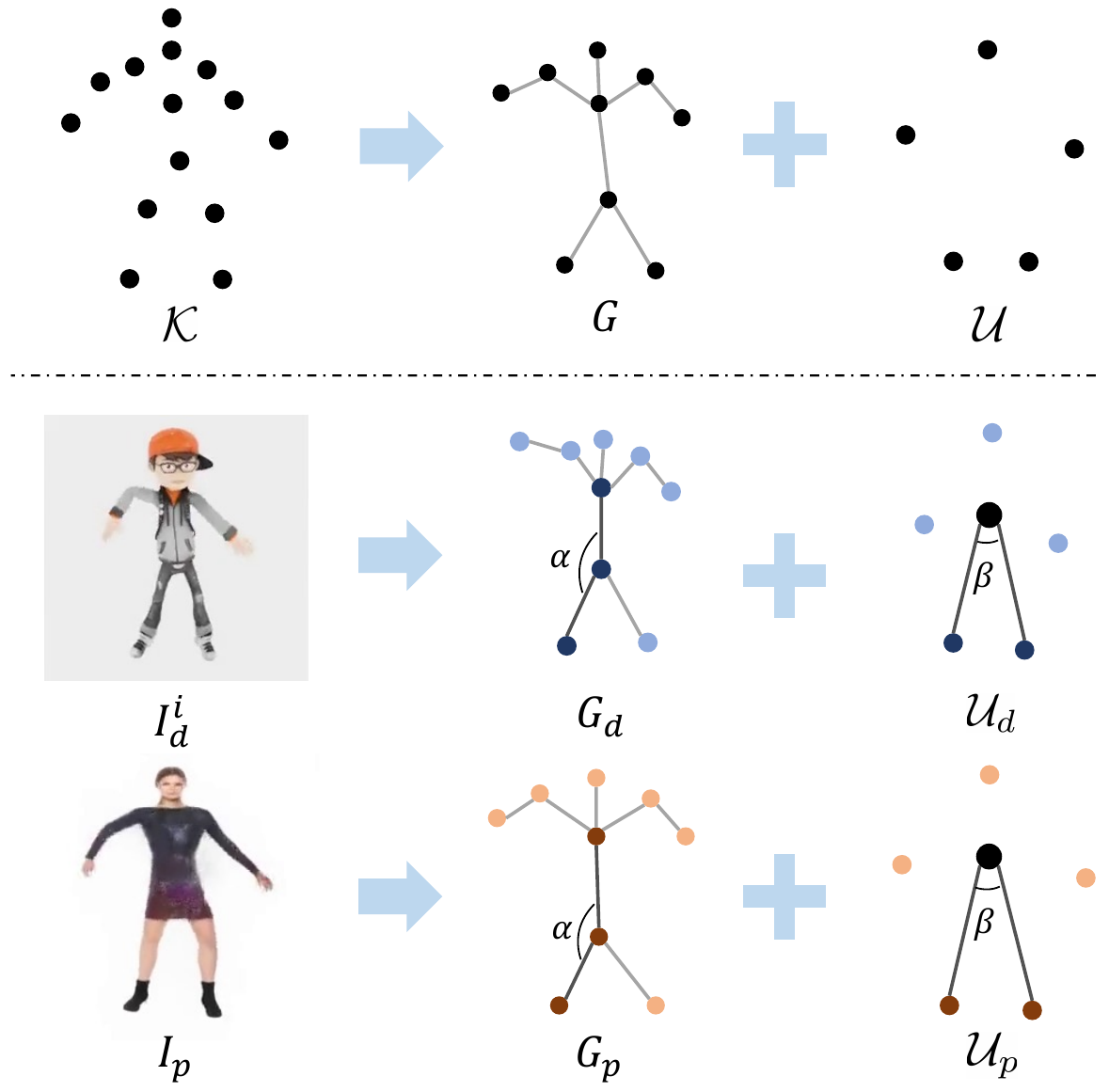}
  \caption{Illustration of our shape-invariant motion adaptation module. The top row show the structure topology, and the bottom two rows represents the motion adaptation stage using structured and unstructured keypoints}
  \label{fig:skeleton}
  % \vspace{-15pt}
\end{figure}

%-------------------------------------------------------------------------
Due to the significant appearance difference between the source image $I_s$ and the driving frame $I_d$, the generated synthesized image $I_p$ often fails to adequately capture the object motion in the driving frame $I_d$. We therefore propose to directly regularize the object pose in $I_p$ with that in $I_d$ based on the extracted motion keypoints. 

However, due to the diversity of the object shapes in $I_p$ and $I_d$, it is undesirable to directly regularize the consistency of their keypoint positions. We therefore propose to discover the intrinsic topology of the object, then regularize the included angles between adjacent object parts of two objects.  

\textbf{Structure Topology Discovery:} To discover the intrinsic object topology, for each driving video, we employ a pre-trained basic MT model to extract the motion keypoints of all frames in the video. Because the motion keypoints roughly describe the objects' moving body parts, two keypoints can be considered to be adjacent if their distance does not change substantially between different frames. 

Formally, given a driving frame $I_d$, we denote its motion keypoints as  $ \mathcal{K}_d = \{{\k_{d}^{i}}|_{i=1}^K\}$, where $K$ is the number of motion keypoints. For each pair of keypoints $\k_{d}^{i}$ and $\k_{d}^j$,  we calculate their $\ell_2$ distance $d_{i,j} = \ell_2(\k_{d}^i, \k_{d}^j)$, where $i \neq j$. The average distance across all frames of all driving videos can then be computed as $\bar{d}_{i,j} = \frac{1}{T}\sum_{t=1}^T d^{(t)}_{i,j}$, where $d^{(t)}_{i,j}$ is the distance in the $t$-th frame, while $T$ is the total number of video frames. Finally, we calculate the total distance diversity of $\k_{d}^{i}$ and $\k_{d}^j$ as follows:
\begin{equation}
  \begin{split}
    &v_{i,j} = \sum_{t=1}^T | d^{(t)}_{i,j} - \bar{d}_{i,j} |
  \end{split}
\end{equation}

Intuitively, the distance diversity describes the stability of the connection between two keypoints. The smaller the distance diversity $v_{i,j}$, the higher the likelihood that the two keypoints will be adjacent. We then use the distance diversities to construct a structure topology graph $G$, where the nodes are keypoints, and the edges are defined according to the distance diversities. Specifically, we define the edge value as follows:
\begin{equation}
    e_{i,j}=
    \begin{cases}
        \frac{(v_{i,j} - \eta)^{2}}{\eta^{2}}, & v_{i,j}<\eta,\\
        0, & \mbox{otherwise}
    \end{cases}
\end{equation}
where $\eta$ is a threshold, and we filter out the edges with high distance diversities, as these imply that the two keypoints are unlikely to be adjacent. Note that the edge value $e_{i,j}$ is within the range of $[0, 1]$. It can be seen as a measurement of the strength of the connection between two keypoints. We will demonstrate that it can also be used as a weight when we regularize the keypoints between driving frame and synthesized image. 

Moreover, it is possible that not all keypoints are connected in a single graph; we select the largest graph as our structure topology graph $G$. We refer to the keypoints in $G$ as the structured keypoints and the others as unstructured keypoints. For improved convenience of presentation, we denote the set of structured keypoints as $\cS$ and their edges as $\cE$, the structure topology graph can be presented as $G = \{\cS, \cE\}$. For unstructured keypoints, we retain only the keypoints and discard their edges, since their connectivities are weak, and denote the set of unstructured keypoints as $\cU$. We present an illustration of the structure topology discovery in the top row of \cref{fig:skeleton}. 

\textbf{Regularizing Structured Keypoints:} 
Given a driving frame $I_d$ and the synthesized $I_p$, we extract their keypoints $\cK_d = \{\k_d\}$ and $\cK_p=\{\k_p\}$ using the basic MT model. To regularize the keypoints in the driving frame $I_d$ and the synthesized $I_p$, we instantiate the structure topology $G$ using the extracted keypoints $\cK_d$ and $\cK_p$, respectively. Taking the driving frame as an example, the instantiated graph is presented as $G_d = \{\cS_d, \cE_d\}$; here, $\cS_d$ is the set of structured keypoints in $I_d$, while $\cE_d$ is the set of corresponding edges which are calculated based on the Euclidean distances between keypoints. The instantiated graph of the synthesized image $G_p = \{\cS_d, \cE_d\}$ can be similarly defined. We illustrate the instantiated graphs in the top of Fig~\ref{fig:skeleton}. 

When examining the structured keypoints, we can observe that considerable differences exist in terms of object shape; this validates our analysis that it is not preferable to directly regularize the keypoint positions. However, the pose can be portrayed as the included angle of each triplet of the connected keypoints in the structure graph. 

Specifically, taking the driving frame as an example, let us define a triplet of connected keypoints as $\t_d = \{\k_d^{i}, \k_d^{j}, \k_d^{k}\}$, where both $\k_d^j$ and $\k_d^k$ are connected to $\k_d^{i}$. We further denote the set of all keypoint triplets in $G_d$ as $\cT_d = \{\t_d^{n}|_{n=1}^N\}$, where $N$ is the total number of triplets. Similarly, we define the set of triplets for the synthesized image as $\cT_p = \{\t_p^{n}|_{n=1}^N\}$. 

For each triplet $\t_d^{n}$ (\resp, $\t_p^{n}$ ), we calculate its included angle and denote it by $\alpha(\t_d^n)$ (\resp, $\alpha(\t_p^n)$). We then regularize the consistency of the corresponding included angles for structured keypoints in the driving frame and the synthesized image as follows:
\begin{equation}
  \begin{split}
    \cL_{rs} = \frac{1}{N}\sum_{n=1}^N \gamma_n| \alpha(\t_{d}^n) - \alpha(\t_{p}^n) |
  \end{split}
\end{equation}
where $\gamma_n$ is the weight for the $n$-th triplet. We calculate $\gamma_n$ using the edge values in the topology graph $G$. Specifically, given any triplet $\t = \{\k^{i}, \k^{j}, \k^{k}\}$ in the topology graph, the weight is computed as $\gamma = e_{i,j}e_{i,k}$. As the edge represents the strength of the connections between two keypoints, it is reasonable to employ the multiplication of the two edges that formed the included angle as the weight for regularization.

\textbf{Regularizing Unstructured Keypoints:} Similarly, given a driving frame $I_d$ and the synthesized $I_p$, we identify their unstructured keypoints $\cU_d$ and $\cU_p$, respectively. Since these unstructured keypoints are disjoint, we constrain them by encoding their included angles with the object centroid. Taking the driving frame as an example, for each pair of keypoints $(\k_d^i, \k_d^j)$ in $\cU_d$, we construct a triplet $\hat{\t}_d = (\k_d^i, \k_d^c, \k_d^j)$ in which $\k_d^c$ is the object centroid, and further denote the included angle as $\beta(\hat{\t}_d)$. We similarly define the corresponding included angle for the synthesized image as $\beta(\hat{\t}_p)$. We then regularize the consistency of the corresponding included angles for the structured keypoints in the driving frame and the synthesized image as follows:
\begin{equation}
  \begin{split}
    \cL_{ru} = \frac{1}{\hat{N}}\sum_{n=1}^{\hat{N}}|\beta(\hat{\t}_{d}^n) - \beta(\hat{\t}_p^n) |
  \end{split}
\end{equation}
where $\hat{N}$ is the number of constructed triplets using structured keypoints in each image. 

Combining the loss of structured and unstructured keypoints, the total loss of our shape-invariant motion adaptation loss can be written as follows:
\begin{equation}
    L_{ma} = L_{rs} + L_{ru}
\end{equation}

%-------------------------------------------------------------------------

\begin{figure}
  \centering
   % \vspace{-15pt}
  \includegraphics[width=0.6\linewidth]{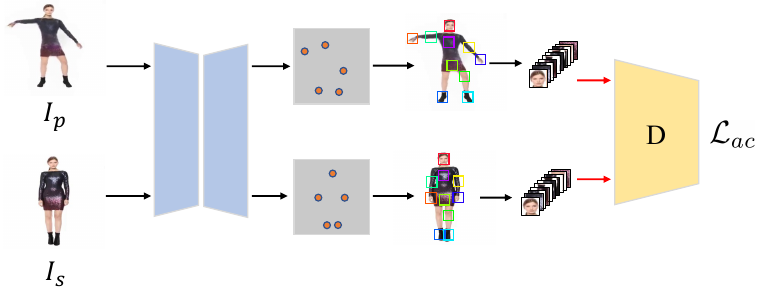}
  \caption{Illustration of our structure-guided appearance consistency module}
  \label{fig:patch}
   % \vspace{-25pt}
\end{figure}

%-------------------------------------------------------------------------

\subsection{Structure-Guided Appearance Consistency Module}
\label{sec:SGAC}
We now consider how the appearance of the synthesized image $I_p$ might be enforced to be similar to that of the source image $I_s$. Note that the object poses in $I_p$ and $I_s$ are different, as we have enforced $I_p$ to mimic the pose of the driving frame. We therefore propose structure-guided appearance consistency module to regularize the appearance consistency of object parts in $I_p$ and $I_c$ to avoid impacting the learned object pose of $I_p$

In particular, we use the predicted motion keypoints to extract image patches of fixed size from both images. After collecting the patches from $ {I}_{p}$ (\resp, $I_s$), a discriminator $\cD$ is then introduced to enforce the appearance consistency between the corresponding patches by means of an adversarial training strategy, as shown in \cref{fig:patch}. The aim of the discriminator is to determine whether the input patches are from ${I}_{p}$  or $I_s$ by minimizing a cross-entropy loss, while the generation model $\cB$ (\ie, the basic MT model) aims at generating pseudo-images $\cB(I_{s}) $, which are difficult to distinguish from the source image $ I_{s}$ by maximizing the cross-entropy loss. Formally, we express the loss of our patch-based appearance consistency module as follows:
\begin{equation}
  \begin{split}
    L_{ac} = & \log \cD(V(I_s)) +\log(1-\cD(V(\cB(I_s))))
  \end{split}
\end{equation}
where $V(\cdot) $ represents the patch extraction operation.

\subsection{Summary}
We combine all losses together to train our proposed MAA model in an end-to-end manner. The overall objective function can be written as follows,
\begin{equation}
    \cL = \cL_{r} + \cL_{c} + \lambda_{ma}\cL_{ma} -  \lambda_{ac}L_{ac}
\end{equation}
where $\lambda_{ma}$ and $\lambda_{ac}$ are tradeoff parameters used to balance the losses. Due to the existence of the discriminator, we optimize the overall loss in an adversarial training manner, \ie, $\min_{\cB}\max_{\cD}\cL$. Detailed training loop is presented in Supplementary materials.

\section{Experiment}
\label{sec:Experiment}

\subsection{Datasets}
% \wenli{Revise the presentation here to avoid confusion.}
% In motion transfer, the model takes a video and an image as input. When considering the cross-domain motion transfer task, the videos and images are from different domain. 
We conduct experiments for two types of object including human body and human face. For the human body animation, we transfer motion from Mixamo-Video to Fashion-Video, and for human face animation we transfer motion from Vox-Celeb to CUHK Face Sketch~(Cufs).

\textbf{Mixamo-Video Dataset} is a synthetic human dancing video dataset newly constructed by ourselves. We collect $15$ characters of 3D human body models and $46$ dancing sequences from the mixamo~\cite{mixamo.com} website, then render dancing videos for these characters and dancing sequences, leading to $ 15\times 46 = 690 $ videos in total with resolution of $ 256\times 256 $. We split ten of the characters as training set and the rest as test set, \ie $460$ and $230$ videos, respectively. Details of the dataset are presented in supplementary materials, and we will release the dataset soon.

\textbf{Fashion-Video Dataset} is a video dataset for showing clothes. It contains $500$ training videos and $100$ testing videos with size of $ 256\times 256 $.Although it is a video dataset, We use it as an image dataset by selecting one frame per video randomly in training stage. 

\textbf{Vox-Celeb} is a video dataset of human talking. It consists of $12,331$ training videos and $444$ testing videos resized to $ 256\times 256 $. 

\textbf{CUHK Face Sketch~(Cufs)} is an image dataset of human face sketches. The dataset contains $305$ images where training set and test set have $250$  and $45$ images \resp. Each image is a sketch drawn by an artist based on a photo taken in a frontal pose with a natural expression. We also resize those images into the size of $ 256\times 256 $.

%-------------------------------------------------------------------------

\begin{table}[]
\centering

% \wenli{Do you have ablation results for MRAA? If not, you can consider to move the ablation results into a separated table, and discuss it in Section 4.4}
% \vspace{-15pt}
\caption{ Quantity results comparison of our method with source only FOMM model and MRAA model. The lower FID and AED values are the better}
% % \vspace{10pt}
\begin{tabular}{c|cc|cc}
\hline
              & \multicolumn{2}{l|}{Mixamo $\longrightarrow$   Fashion} & \multicolumn{2}{l}{Vox $\longrightarrow$ Cufs} \\
              & FID $\downarrow$                          & AED $\downarrow$                         & FID $\downarrow$                     & AED $\downarrow$                     \\ \hline
MRAA   & 177.3                        & 0.376                       & 127.1                   & 0.764                   \\ \hline
FOMM   & 175.9                        & 0.359                       & 112.5                   & 0.693                   \\ \hline
Ours~(MRAA)    & 72.1                         & 0.289                       & 86.5                       & 0.627                       \\ \hline
Ours~(FOMM)          & \textbf{61.7}                         & \textbf{0.274}                       & \textbf{50.1}                    & \textbf{0.573}                     \\ \hline
\end{tabular}

\label{tab:FID-AED}
% \vspace{-15pt}
\end{table}

%-------------------------------------------------------------------------
% % \vspace{-10pt}

\subsection{Quantitative Results}
\textbf{Metrics:} As the ground-truth video are not available in cross-domain motion transfer, to quantitatively assess the synthesized videos, we employ two metrics for evaluate generative models as follows
\begin{itemize}
    \item \textbf{Fr\'{e}chet Inception distance (FID)\cite{heusel2017gans}} This score indicates the overall quality of generated frames, it compares the feature statistics of generated frames and real images, then calculates the distance between them.
    \item \textbf{Average Euclidean Distance (AED)}\cite{siarohin2019first} Considering the generated images share the same identity with source images, we utilize AED to evaluate the identity similarity between them. It also computes the feature distance between two input images. Specifically, a pre-trained person re-identification network~\cite{hermans2017defense} and a pre-trained facial identification network~\cite{amos2016openface} are used to extract identity feature representations for human body and human face dataset, respectively.
\end{itemize}

\textbf{Results:}
As aforementioned, unsupervised motion transfer models like FOMM\cite{siarohin2019first} or MRAA\cite{siarohin2021motion} can be integrated into our MAA framework as the basic motion transfer model. We conduct experiments by respectively using FOMM and MRAA as our basic motion transfer model, and take the original FOMM and MRAA as the corresponding baseline for comparison. For both methods, the baseline models are trained on the driving video dataset without considering the cross-domain issue. Note that the newly proposed modules in our MAA model are only used in training stage, and the model in the test stage share the same architecture with the baseline FOMM or MRAA model. 

We report the results for Mixamo-Video $ \rightarrow $ Fashion-Video and Vox $ \rightarrow $ Cufs in~\cref{tab:FID-AED}. Comparing with the FOMM and MRAA model, our proposed MAA approach  achieves a much better performance. In particular, compared with FOMM, we achieve a FID of $61.7$ and an AED of $0.274$ for  Mixamo-Video $ \rightarrow $ Fashion-Video, and $50.1$ \vs  $112.5$ and $0.573$ \vs $0.693$ for Vox $ \rightarrow $ Cufs, respectively. Compared with MRAA, we achieve a FID of $72.1$ and an AED of $0.289$ for  Mixamo-Video $ \rightarrow $ Fashion-Video, and $86.5$ \vs  $127.1$ and $0.627$ \vs $0.764$ for Vox $ \rightarrow $ Cufs, respectively. Note that, for both FID and AED metrics, smaller value is better. The large improvement indicates that the cross-domain motion transfer is challenging for the traditional FOMM and MRAA method, while our MAA model works well on the cross-domain scenario. We observe that MRAA performs worse than FOMM in the cross-domain motion transfer task, although the previous work shows MRAA usually performs better than FOMM in the traditional single-domain motion transfer\cite{siarohin2021motion}. This possibly dues to that the PCA based motion estimation in the MRAA method is non-parametric and less flexible for cross-domain motion transfer.

%-------------------------------------------------------------------------

\begin{figure}
  \centering
  \includegraphics[width=0.9\linewidth]{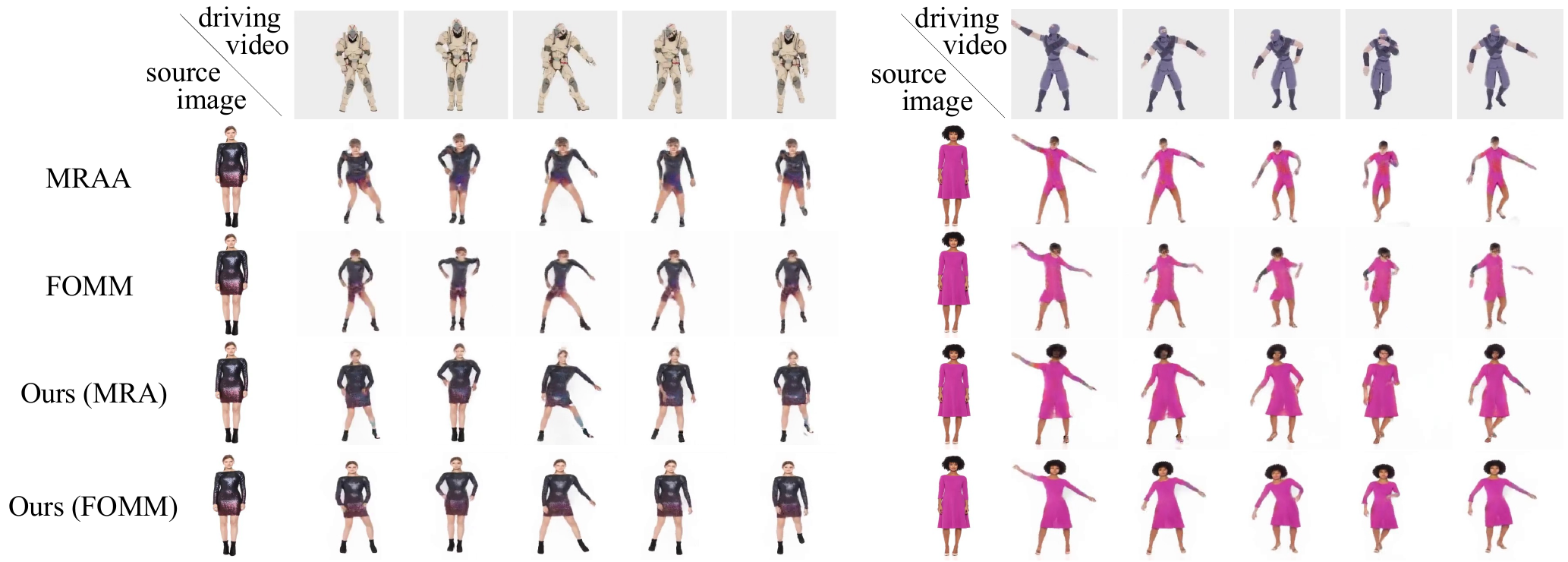}
  \caption{Visualization results of FOMM,MRAA and ours method on human body datasets}
  \label{fig:body-result}
\end{figure}

%-------------------------------------------------------------------------

%-------------------------------------------------------------------------

\begin{figure}
  \centering
  \includegraphics[width=0.9\linewidth]{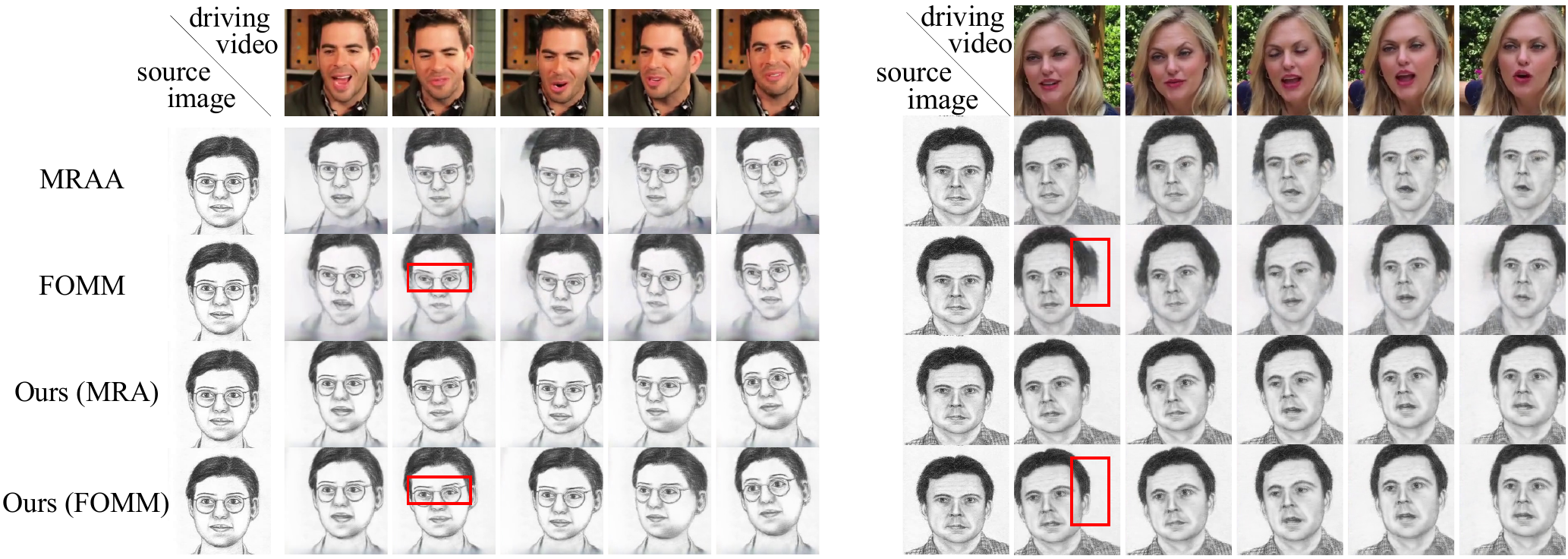}
  \caption{Visualization results of FOMM,MRAA and ours method on human face datasets}
  \label{fig:face-result}
\end{figure}

%-------------------------------------------------------------------------

\subsection{Qualitative Results}
We visualize the generated results to gain an intuitive assessment of FOMM, MRAA and our MAA models for cross-domain human body and human face animation in \cref{fig:body-result} and \cref{fig:face-result}, respectively.  In each figure, two pairs of results are visualized in the left and right parts, respectively. Driving frames extracted from the test video are displayed on the top row, while the source images are showed at the most left column of each part. 

For human body animation, as shown in \cref{fig:body-result}, the results generated by the FOMM and MRAA model obviously suffer from domain shift problem. Although the motion of driving video is roughly captured, the human body shape of source image is rarely preserved, and notable artifacts can be observed in almost all frames of the synthesized video. In contrast, our MAA model is able to capture the motion of the driving frames while properly preserving the appearance of the source image.

For human face animation, as shown in \cref{fig:face-result}, the FOMM and MRAA model could generate results with a rough motion of driving frames and a similar facial appearance with source image. However, the quality of synthesized image are not satisfactory where artifacts are obvious to observe. For example, artifacts on glasses and heads can be observed for FOMM results as highlighted in the red bounding boxes. These differences in qualitative results clearly demonstrate the effectiveness of our proposed MAA model for cross-domain motion transfer.

\subsection{Ablation Study}

To study the impact of our proposed modules, we further conduct ablation study on both human body and human face datasets. The FOMM is used as the basic motion transfer model. The quantitative results are shown in \cref{tab:ablation}, where 'w/o CYC', 'w/o SIMA' and 'w/o SGAC' means removing the cyclic training pipeline, shape-invariant motion adaptation and structure-guided appearance consistency of FOMM model, respectively.

%-------------------------------------------------------------------------

\begin{table}[]
\centering
\caption{Ablation results comparison of FOMM and our ablated models.}
\begin{tabular}{c|cc|cc}
\hline
              & \multicolumn{2}{l|}{Mixamo $\longrightarrow$   Fashion} & \multicolumn{2}{l}{Vox $\longrightarrow$ Cufs} \\
              & FID $\downarrow$                          & AED $\downarrow$                         & FID $\downarrow$                     & AED $\downarrow$                     \\ \hline
FOMM   & 175.9                        & 0.359                       & 112.5                   & 0.693                   \\ \hline
w/o CYC   & 136.9                        & 0.354                       & 74.1                    & 0.633                   \\ \hline
w/o SIMA    & 80.2                         & 0.303                       & 60.7                    & 0.622                   \\ \hline
w/o SGAC    & 67.7                         & 0.284                       & 55.2                       & 0.603                       \\ \hline
Ours~(FOMM)          & \textbf{61.7}                         & \textbf{0.274}                       & \textbf{50.1}                    & \textbf{0.573}                     \\ \hline
\end{tabular}

\label{tab:ablation}
\end{table}

%-------------------------------------------------------------------------

\begin{figure}
  \centering
  \includegraphics[width=0.45\linewidth]{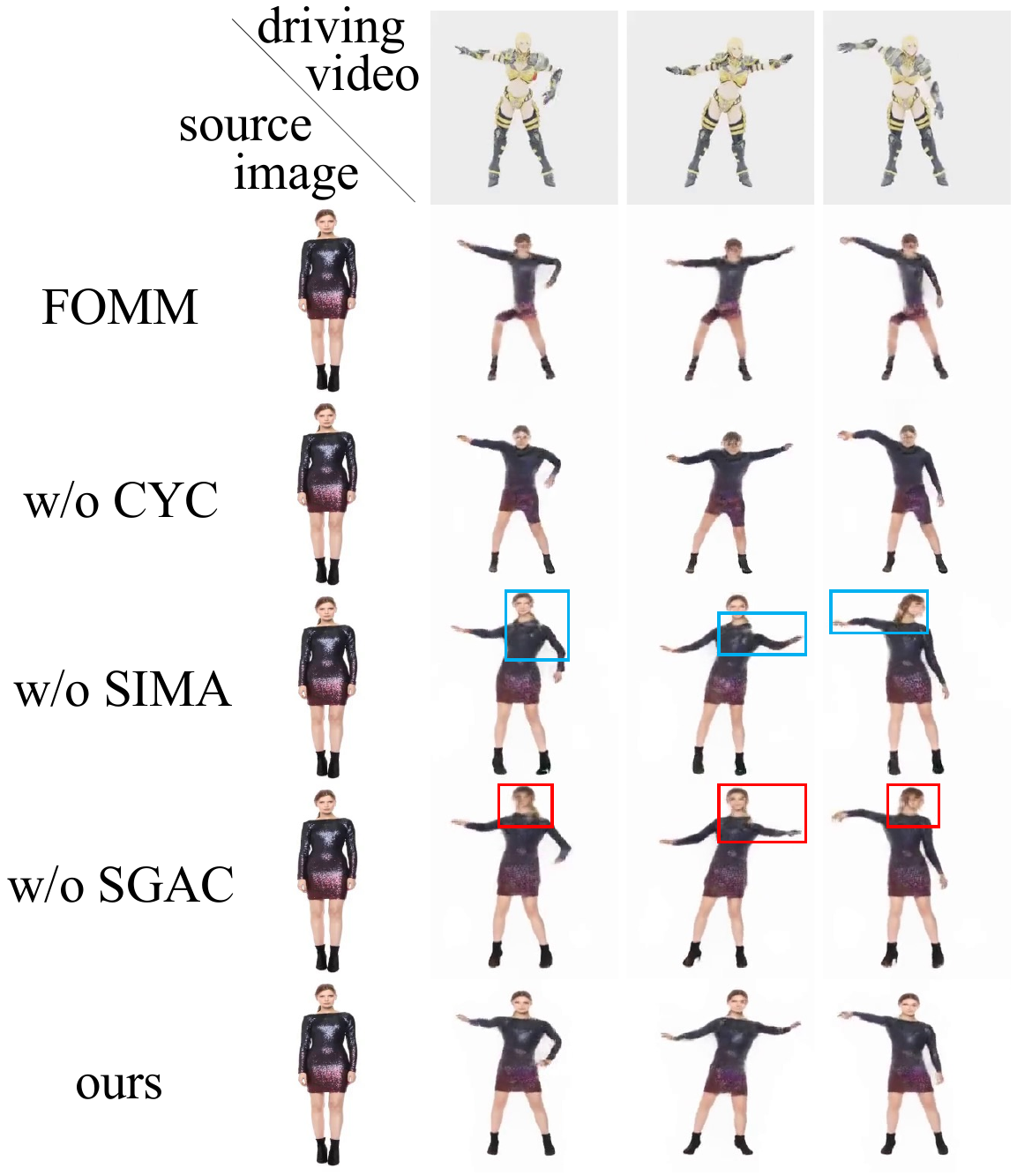}
  \caption{Visualized ablation study results on the human body datasets}
  \label{fig:ablation}
\end{figure}

For both human body and human face animation, as shown in \cref{tab:ablation}, we observe considerable performance drops on both AED and FID for w/o CYC, which again confirms the importance to explicitly consider the cross-domain issue when performing motion transfer across domains. Similar observations can be obtained on human face dataset, which is also confirmed in the qualitative results in \cref{fig:ablation}. Other ablation settings w/o SIMA and w/o SGAC also degrade the performance considerably, which validates the necessity of using the two modules for generating satisfactory synthesized video in cross-domain motion transfer.

To show the effect of each module intuitively, we further visualize the synthesized results in \cref{fig:ablation}. We observe that the result of w/o CYC has richer details than that of FOMM model. For example, the face and the clothes are clearer. However, compared with our final MAA result, it still drops important motion and appearance information. Moreover, we observe the result of w/o SIMA are able to preserve relative rich appearance information, however, the pose of driving frames are not transferred properly without the help of motion consistency module. For example, artifacts can be observed for the poses of the arms and heads as highlighted in the blue rectangles. And, on the third row, w/o~SGAC performs well in pose transferring but fails to preserve source image appearance without SGAC, especially for the details of human face as highlighted in red rectangles. These observations confirm the effectiveness of the modules proposed in our MAA approach.

%-------------------------------------------------------------------------

%-------------------------------------------------------------------------

\subsection{User Study}
To further evaluate our model, we additionally conduct a user study. In particular, we randomly select 50 pairs of source domain driving videos and target domain source images for both human body animation and human face animation, and generate result videos in an ablation setting. For each dataset, we compare results of our final MAA model with those of FOMM and three ablation methods, respectively. The comparison are evaluated by 25 users according to three aspects, motion, appearance~(denoted as app. in \cref{tab:User-Study}) and overall, respectively. 

The user preferences are shown in \cref{tab:User-Study}. We observe that all scores are above $0.5$, which means our results are preferred by the majority of users for all aspects in all settings. For motion aspect, fewer users prefer w/o SIMA than other settings when compared with MAA model on both datasets ($0.748$ vs. $0.717$ and $0.679$ for human body, and $0.571$ vs. $0.704$ for human face), which indicates SIMA improves the motion of generated results. For appearance aspect, fewer user prefer w/o SGAC than other ablation settings when compared with MAA model in human body dataset ($0.715$ vs. $0.711$ and $0.699$), which indicates SGAC contributes to appearance of generated results.

%-------------------------------------------------------------------------

\begin{table}[]
\centering
\caption{User study results. We compare the Ours (FOMM) model to every ablation model, and the values represent the user preferences to Ours (FOMM) model}
% \resizebox{\textwidth}{10mm}{
\begin{tabular}{c|ccc|ccc}
\hline
            & \multicolumn{3}{c|}{Mixamo $\longrightarrow$ Fashion} & \multicolumn{3}{c}{Vox $\longrightarrow$ Cufs} \\
            & motion          & appearance           & overall         & motion        & appearance        & overall       \\ \hline
FOMM & 0.888           & 0.983          & 0.978           & 0.845         & 0.792       & 0.875         \\ \hline
w/o CYC  & 0.717           & 0.699          & 0.732           & 0.571         & 0.615       & 0.626         \\ \hline
w/o SIMA  & 0.748           & 0.711          & 0.702           & 0.704         & 0.655       & 0.675         \\ \hline
w/o SGAC   & 0.679           & 0.715          & 0.725           & 0.593            & 0.617          & 0.575            \\ \hline
\end{tabular}

\label{tab:User-Study}
\end{table}

%-------------------------------------------------------------------------

\section{Conclusion}
\label{sec:Conclusion}
In this paper, we propose a Motion and Appearance Adaptation~(MAA) approach for cross-domain motion transfer. In MAA, we design a shape-invariant motion adaptation module to enforce the consistency of the angles of object parts in two images to capture the motion information. Meanwhile, we introduce a structure-guided appearance consistency module to regularize the similarity between the patches of the synthesized image and the source image. The experimental results demonstrates the effectiveness of our proposed method. 

\begin{FlushLeft}\textbf{Acknowledgement}. This work is supported by the Major Project for New Generation of AI under Grant No. 2018AAA0100400, the National Natural Science Foundation of China (Grant No. 62176047), Sichuan Science and Technology Program (No. 2021YFS0374, 2022YFS0600), Beijing Natural Science Foundation (Z190023), and Alibaba Group through Alibaba Innovation Research Program. This work is also partially supported by the Science and Technology on Electronic Information Control Laboratory.\end{FlushLeft}

\bibliographystyle{splncs04}
\bibliography{egbib}
\end{document}